\documentclass[11pt]{TinLatex} 
\pdfoutput=1

\usepackage{graphicx}
\usepackage{epstopdf}

\usepackage{amsmath,amsxtra,amssymb,latexsym, amscd}

\usepackage[mathscr]{eucal}

\usepackage{hyperref}
\hypersetup{
	colorlinks=false,
	pdfborder={0 0 0}
}

\usepackage{xspace}

\def\centerbmp#1#2#3{\vskip#2\relax\centerline{\hbox to#1{\special
  {bmp:#3 x=#1, y=#2}\hfil}}}

\def\centereps#1#2#3{\vskip#2\relax\centerline{\hbox to#1{\special
  {eps:#3 x=#1, y=#2}\hfil}}}

\def\centerwmf#1#2#3{\vskip#2\relax\centerline{\hbox to#1{\special
  {wmf:#3 x=#1, y=#2}\hfil}}}

\def\centerps#1#2#3{\vskip#2\relax\centerline{\hbox to#1{\special
  {ps:#3 x=#1, y=#2}\hfil}}}

\input dih.tex

\font\it=cmti12 at 11pt
\font\bf=cmbx12 at 11pt

\advance\hoffset0.4cm
\font\bfh=cmssbx10 scaled 1400
\font\bfl=cmssbx10 scaled \magstep1

\font\cn=cmr10

\def\ed{\end{document}}
\def\vuong{\raise-0.16cm\hbox{$^\blacksquare$}}

\makeatletter
\DeclareRobustCommand\onedot{\futurelet\@let@token\@onedot}
\def\@onedot{\ifx\@let@token.\else.\null\fi\xspace}

\makeatother

\usepackage{caption, array}
\usepackage{adjustbox}
\usepackage{amsmath}
\DeclareCaptionLabelFormat{tablel}{#1 #2}
\usepackage{minibox}
\usepackage{etoolbox}
\usepackage{amsmath}
\usepackage{enumitem}
\usepackage{amssymb,amsmath}
\usepackage{xcolor}
\usepackage{comment}
\usepackage{pgfplots}
\usepackage{url}
\usepackage[vietnamese,english]{babel}
\usepackage{amsmath}
\usepackage{multirow}
\usepackage{color}
\usepackage{subcaption}

\renewcommand{\thetable}{\arabic{table}}
\begin{document}
\makeatletter	 
\setcounter{page}{1}
\renewcommand{\ps@plain}{\renewcommand{\@oddhead}{\hfill\begin{tabular}{r}
\small{\emph{Journal of Computer Science and Cybernetics, V.xx, N.x (20xx), 1--}} \\
\footnotesize{DOI no. 10.15625/1813-9663/xx/x/xxxx}
\end{tabular}
}

    \renewcommand{\@evenhead}{\@oddhead}    \renewcommand{\@oddfoot}{\small\hfill{\copyright\ 20xx Vietnam  Academy of Science \& Technology}}    \renewcommand{\@evenfoot}{\@oddfoot}}
    \makeatother

\title{Neural Machine Translation between Vietnamese and English: an Empirical Study}
\author{
	{\cn Hong-Hai Phan-Vu\it $^1$, Viet-Trung Tran\it $^{1,3}$, Van-Nam Nguyen\it $^2$, Hoang-Vu Dang\it $^2$, Phan-Thuan Do\it $^1$}
	\vskip.5cm
	{\it $^1$Hanoi University of Science and Technology (HUST)}\\
	{\it $^2$FPT Technology Research Institute, FPT University}\\
	{\it $^3$Corresponding author. Email: \href{mailto:trungtv@soict.hust.edu.vn}{trungtv@soict.hust.edu.vn}}
			}
\maketitle
\renewcommand\refname{\normalsize \centerline{ REFERENCES}}
\pagestyle{plain}
\pagestyle{myheadings}

\begin{abstract}
Machine translation is shifting to an end-to-end approach based on deep neural networks. The state of the art achieves impressive results for popular language pairs such as English - French or English - Chinese. However for English - Vietnamese the shortage of parallel corpora and expensive hyper-parameter search present practical challenges to neural-based approaches. This paper highlights our efforts on improving English-Vietnamese translations in two directions: (1) Building the largest open Vietnamese - English corpus to date, and (2) Extensive experiments with the latest neural models to achieve the highest BLEU scores. Our experiments provide practical examples of effectively employing different neural machine translation models with low-resource language pairs.
\end{abstract}

\section{Introduction}
\label{sec:intro}
Machine translation is shifting to an end-to-end approach based on deep neural networks. Recent studies in neural machine translation (NMT) such as \cite{Vaswani:2017, Bahdanau:2014, Wu:2016,Gehring:2017} have produced impressive advancements over phrase-based systems while eliminating the need for hand-engineered features. Most NMT systems are based on the encoder-decoder architecture which consists of two neural networks. The encoder compresses the source sequences into a real-valued vector, which is consumed by the decoder to generate the target sequences. 
The process is done in an end-to-end fashion, demonstrated the capability of learning representation directly from the training data. 

The typical sequence-to-sequence machine translation model consists of two recurrent neural networks (RNNs) and an attention mechanism \cite{Bahdanau:2014, Luong:2015b}. Despite great improvements over traditional models \cite{Wu:2016, Sennrich:2016, Luong:2015a}  this architecture has certain shortcomings, namely that the recurrent networks are not easily parallelized and limited gradient flow while training deep models.

Recent designs such as ConvS2S \cite{Gehring:2017} and Transformer \cite{Vaswani:2017} can be better parallelized while producing better results on WMT datasets. However, NMT models take a long time to train and include many hyper-parameters. There is a number of works that tackle the problem of hyper-parameter selection \cite{Britz:2017, Popel:2018} but they mostly focus on high-resource language pairs data, thus their findings may not translate well to low-resource translation tasks such as English-Vietnamese. Unlike in Computer Vision \cite{Huang:2016, Kandaswamy:2014}, the task of adapting parameters spaces from one NMT model to other NMT models is nearly impossible \cite{Britz:2017}. This reason limits researchers and engineers to reach good-chose hyper-parameters and well-trained models.

To date there are several research works on English-Vietnamese machine translation such as \cite{Le:2003, Dinh:2003, Ho:2009, Nguyen:2016}, using traditional methods with modest BLEU scores. Some newer works such as \cite{Luong:2015c, Huang:2017} experimented on the IWSLT English-Vietnamese dataset \cite{IWSLT:2015} and showed great potential to improve English-Vietnamese translation tasks using more data and more complex models. 

In \cite{Phan:2017} the authors introduced datasets for bilingual English-Vietnamese translation and attained state-of-the-art BLEU scores using sequence-to-sequence models and vanilla preprocessing. In this work we perform extensive experiments on large-scale English-Vietnamese datasets with the latest NMT architectures for further improvements in BLEU scores and report our empirical findings. 

Our main contributions are as follows: (1) A brief survey of current state of the art in NMT. (2) The construction of a large parallel corpus for English-Vietnamese translation, which will be publicly available. (3) Implementation and experimentation of the newest models, and our source code will also be shared. (4) Empirical findings on tuning the aforementioned models.

\section{Latest NMT architectures}

\subsection{Sequence-to-Sequence RNNs}
Here we introduce the sequence-to-sequence model based on an encoder-decoder architecture with attention mechanism \cite{Luong:2015b}. Let $(X, Y)$ be the pair of source and target sentences, where $X = x_1, \ldots, x_m$ is a sequence of $m$ symbols and $Y=y_1, \ldots, y_n$ a sequence of $n$ symbols. The encoder function $f_{enc}$ maps the input sequence $X$ to a fixed size vector, which the decoder function $f_{dec}$ uses to generate the output sequence $Y$.

While $f_{dec}$ is usually a uni-directional RNN, $f_{enc}$ can be a uni-directional, bi-directional or hybrid RNN. In this work we consider bi-directional encoders. Each state of $f_{enc}$ has the form $\overline{h}_i = [\overrightarrow{h_i},\overleftarrow{h_i}]$ where the components encode $X$ in forward and backward directions. The auto-regressive decoder $f_{dec}$ then predicts each output token $y_i$ from the recurrent state $s_i$, the previous tokens $y_{<i}$ and a context vector $c_i$. 

The context vector $c_i$ is also called attention vector and depends on encoder states together with the current decoder state. Among known attention architectures, in this work we use the most efficient as described in \cite{Luong:2015b}.  At the decoding step $t$, an alignment vector $a_t$ is derived from the current decoder hidden state $h_t$ and each encoder hidden state $\overline{h}_s$. The context vector $c_t$ is a weighted average over all encoder states with weights $a_t$.
\begin{align}
a_t(s) &= align(h_t, \overline{h}_s) \\
c_t &= \sum a_th_s
\end{align}

The context vector $c_t$ is concatenated with the current hidden decoder state $h_t$ to produce an attentional state $\tilde{h}$, which is fed through a softmax layer to produce the predicted distribution.

\begin{align}
\tilde{h} &= tanh(W_c[c_t; h_t]) \\
p(y_t|y_{<t}, x) &= softmax(W_s\tilde{h}_s)
\end{align}

\subsection{The Convolutional Sequence-to-Sequence Model}
\begin{figure}[h!]
\begin{minipage}{0.48\textwidth}
\centering
\includegraphics[width=0.94\textwidth]{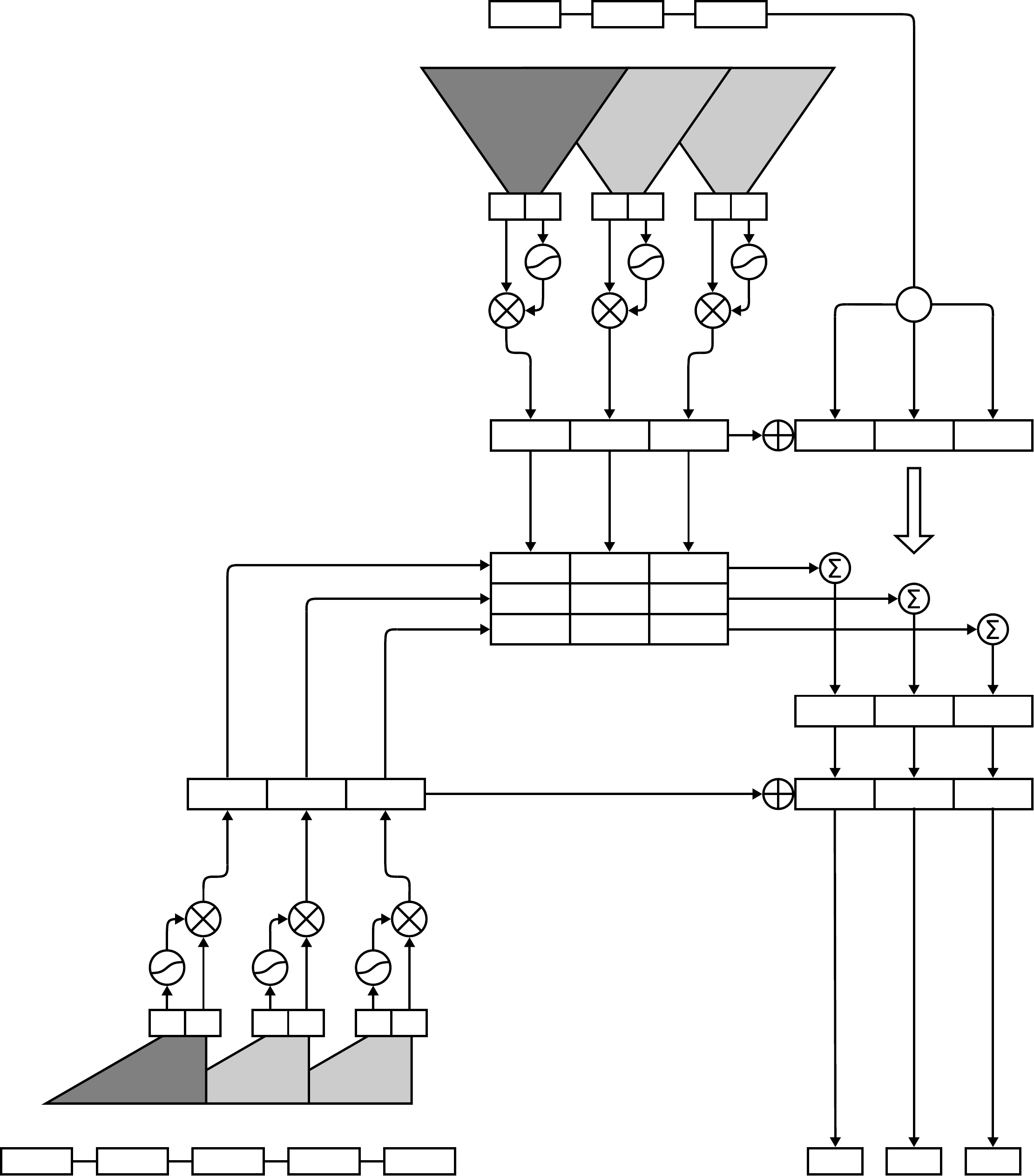}
\caption{\small The Convolution Sequence-to-Sequence model architecture, adapted from \cite{Gehring:2017}}
\label{convs2s_overal}
\end{minipage}~
\begin{minipage}{0.48\textwidth}
\centering
\includegraphics[width=0.94\textwidth]{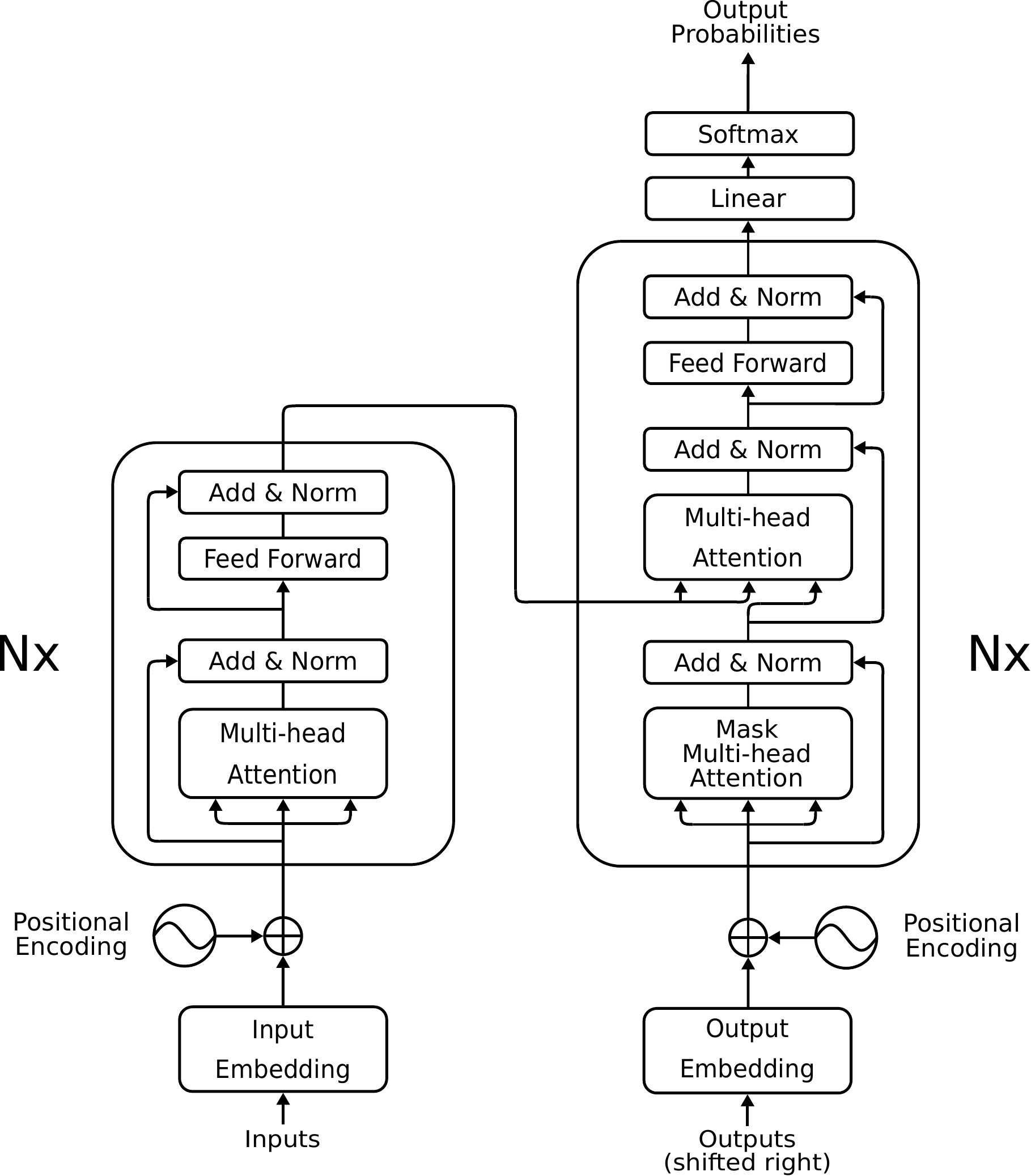}
\caption{\small Overall architecture of the Transformer}
\label{tf_overal}
\end{minipage}
\end{figure}
The Convolutional Sequence-to-Sequence Model (ConvS2S) \cite{Gehring:2017} is a sequence-to-sequence model that uses a fully convolutional architecture. The model is equipped with gated linear units \cite{Dauphin:2016} and residual connections \cite{He:2015}.

\subsubsection{Position Embeddings}
Because the CNN itself can not convey positional information, ConvS2S uses position embeddings to tackle this problem. The input element $x=(x_1,\ldots, x_m)$ is represented as  a vector $z = w + p$ where $w = (w_1,\ldots, w_m)$ embeds the symbols $x_i$ into an Euclidean space $ \mathbb{R}^{f}$ and $p=(p_1, \ldots, p_m)$ embeds the positions of the $x_i$ into $\mathbb{R}^{f}$. The same process is applied to the output elements generated by the decoder network, and the resulting representations are fed back into the decoder.

\subsubsection{Convolutional Layer Structure}
We denote the output of the $i^{th}$ layer by $e^{i} = (e_1^i, \ldots, e_n^i)$ for the encoder network and $d^{i} = (d_1^i, \ldots, d_o^i)$ for the decoder network. In the model, each layer contains a one dimensional convolution followed by a non-linearity. 
Each convolution kernel is parameterized as a weight $W \in \mathbb{R}^{2s \times ks}$ and a bias $b_w \in \mathbb{R}^{2s}$. The kernel's input is a matrix $X \in \mathbb{R}^{k\times s}$ which is a concatenation of $k$ input elements embedded in $s$ dimensions, the kernel's output is a vector $Y \in \mathbb{R}^{2s}$ that has twice the dimensionality of the input elements. Each group of $k$ output elements of the previous layer are operated by a subsequence layer. The non-linearity is the gated linear unit (GLU:\cite{Dauphin:2016}) which implements a gating mechanism over the output of the convolution $Y = [A\ B] \in \mathbb{R}^{2s}$:
\begin{equation}
v([A\ B]) = A \otimes \sigma(B)
\end{equation}
where $A, B \in \mathbb{R}^{s}$ are the non-linearity input, $\otimes$ is the point-wise multiplication and the output $v([A\ B]) \in \mathbb{R}^s$ has half size of $Y$. The gates $\sigma(B)$ control which inputs A of the current context are relevant \cite{Gehring:2017}.
 
Residual connections from the input of each convolution to the output are applied, similar to \cite{He:2015}
\begin{equation}
d^i_j = v(W^i[d^{i-1}_{ j-k/2}, \ldots, d^{i-1}_{j+k/2}] + b^i_w) + d_j^{i-1}
\end{equation}

The convolution outputs that are of size $2s$ are mapped to the embedding of size $f$ by linear projections. These linear mappings are applied to $w$ while feeding embeddings to the encoder network, to the encoder output $e^i_j$, to the final layer of the decoder just before the $softmax$ $d^L$ and to all decoder layers $d^i$ before computing the scores the attentions.

Finally, a distribution over the $T$ possible next target elements $y_{j+1}$ is computed by transforming the top decoder output $d^L_j$ via a linear layer with weights $W_o$ and bias $b_o$:
\begin{equation}
p(y_{j+1} | y_1,\ldots, y_j, x) = softmax(W_od_j^L + b_o) \in \mathbb{R} ^ T
\end{equation}

\subsubsection{Multi-step Attention}
In ConvS2S, the attention mechanism is applied separately for each encoder layer. The attention mechanism works as multiple ``hops'' \cite{Sukhbaatar:2015} compared to single step attention \cite{Bahdanau:2014}, \cite{Luong:2015b}, \cite{Zhou:2016}, \cite{Wu:2016}. At the decoder layer $i$, the attention $a_{kj}^i$ of state $k$ and the source element $j$ are computed as a dot-product between the decoder state summary $v^i_k$ and each output $e_j^u$ of the last encoder layer $u$:
\begin{equation}
a^i_{kj} = \frac{\exp(v_k^i \cdot e_j^u)}{\sum_{t=1}^m\exp(v^i_k \cdot e_j^u)}
\end{equation}
where $v_k^i$ is combined of the current decoder state $d_k^i$ and the embedding of the previous target element $g_k$:
\begin{equation}
v^i_k = W_v^i d_k^i + b_v^i + g_k
\end{equation}

The conditional input vector $c_k^i$ to the current decoder layer is a weighted sum of the encoder output as well as the input embeddings $z_j$:
\begin{equation}
c^i_k = \sum_{j=1}^m a_{kj}^i(e_j^u + z_j)
\end{equation}
After that, $c_k^i$ is added to the output of the corresponding decoder layer $d_k^i$. This attention mechanism can be seen as determining useful information from the current layer to feed to the subsequent layer. The decoder can easily access the attention history of $k-1$ previous time steps. Therefore, the model can take into account which previous inputs have been attended more easily than recurrent networks \cite{Gehring:2017}.

\subsection{The Transformer Model}
Unlike other transduction models, Transformer does not use RNNs or CNNs for modeling sequences. It has been claimed by authors to be the first transduction model to rely entirely on self-attention to compute representations of its input and output \cite{Vaswani:2017}. Like other competitive sequence transduction models, Transformer has an encoder and a decoder. The model is auto-regressive, consuming at each step the previous generated symbols as additional input to emit the next symbol. Compared to RNNs the proposed self-attention mechanism allows for a high degree of parallelization in training, while relying on positional embeddings to capture global dependencies within each sequence.
\subsubsection{Overall Structure}
Like ByteNet \cite{Kalchbrenner:2016} or ConvS2S \cite{Gehring:2017}, the decoder is stacked directly on top of the encoder.  Without the recurrence or the convolution, Transformer encodes the positional information of each input token by a \textit{position encoding} function. Thus the input of the bottom layer for each network can be expressed as $Input = Embedding + Positional Encoding$.

The encoder has several identical layers stacked together. Each layer consists of a \textit{multi-head self-attention mechanism} and a  \textit{position-wise fully connected feed-forward network}. Each of these sub-layers has a residual connection around itself, followed by layer normalization \cite{LeiBa:2016} (Figure \ref{tf_overal}).  The output of each sub-layer is $LayerNorm(x + Sublayer(x))$ where $x$ is the sub-layer input and $Sublayer$ is the function implemented by the sub-layer itself. The outputs of all sub-layers and the embedding layers in the model are vectors of dimension $d_{model}$.

The decoder is also a stack of identical layers, each layer comprising three sub-layers. At the bottom is a masked multi-head self-attention, which ensures that the predictions for position $i$ depend only on the known outputs at the positions less than $i$. In the middle is another multi-head attention which performs the attention over the the encoder output. The top of the stack is a position-wise fully connected feed-forward sub-layer.

The decoder output finally goes through a linear transform with softmax activation to produce the output probabilities. The final linear transform shares the same weight matrix with the embedding layers of the encoder and decoder networks, except that the embedding weights are multiplied by $\sqrt{d_{model}}$.

\subsubsection{Attention}
The attention is crucial in NMT. It maps a \textit{query} and a set of \textit{key-value} pairs to an output. The output of the attention is a weighted sum of the \textit{values} whose weights show the correlation between each \textit{key} and \textit{query}. The novelty is that the Transformer's attention is a \textit{multi-head self-attention}. In the Transformer's architecture, the $query$ is the decoder's hidden state, the $key$ is the encoder's hidden state and the $value$ is the normalized weight measuring the ``attention'' that each \textit{key} is given. It is assumed that the queries and the keys are of dimension $d_k$ and the values are of dimension $d_v$.

\begin{itemize}
\item Scaled dot-product attention:
Let $Q$ be the matrix of queries, $K$ be the matrix of keys and $V$ be the matrix of values. The attention is calculated as follows:
\begin{equation}
Attention(Q, K, V) = softmax(\frac{QK^T}{\sqrt{d_k}})V.
\end{equation}

Instead of using a single attention function, Transformer uses multi-head attentions. The multi-head attention consists of $h$ layers (heads). The queries, keys and values are linearly projected to $d_k$ and $d_v$ dimensions.  Each head receives a set of projections and performs a separate attention function yielding $d_v$-dimensional output values. The heads' outputs are concatenated and projected, resulting in the final multi-head attention output.
\begin{equation}
MultiHead(Q, K, V) = Concat(head_{1}, head_{2},\dots, head_{h})W^{O}
\end{equation}
where $head_{i} = Attention(QW_{i}^{Q}, KW_{i}^{K}, VW_{i}^{K})$.
The projections are parameter matrices $W_i^Q \in \mathbb{R}^{d_{model} \times d_k}$, $W_i^K \in \mathbb{R}^{d_{model} \times d_k}$, $ W_{i}^{V} \in \mathbb{R}^{d_{model} \times d_v}$ and $W_{i}^{O} \in \mathbb{R}^{hd_{v} \times d_{model}}$.
\end{itemize}
If we set $d_k = d_v = d_{model}/h$, the multi-head attention then has the same computational cost as a single full-dimensionality attention.
The Transformer's attention mechanism imitates the  classical attention mechanism where the attention queries are previous decoder layer outputs, the keys and the values (memory) are the encoder layer outputs.

\subsubsection{Position-wise Feed-forward Networks}
The fully connected feed-forward network (FFN) at the top of each layer is applied to each input position separately and identically. Each FFN here consists of two linear transformations with a ReLU activation in between, acting like a stack two convolutions with kernel size 1:
\begin{equation}
FFN(x) = ReLU(xW_1 + b_1)W_x + b_2.
\end{equation}
\subsubsection{Positional Encoding}
There are many types of positional encodings, including both learned and fixed variants \cite{Gehring:2017}. Here the positional encodings are chosen as follows:
\begin{align}
PE(pos, 2i) &= sin(\frac{pos}{10000^{2i/d_{model}}}) \\
PE(pos, 2i+1) &= cos(\frac{pos}{1000^{2i/d_{model}}}),
\end{align}
where $pos$ is the position and $i$ is the dimension. The authors hypothesized this function would allow the model to easily learn to attend by relative positions \cite{Vaswani:2017}. Their experiments showed that these encodings have the same performance as learned positional embedding. Furthermore they allow the model to extrapolate to sequences longer than the training sequences.

\section{Parallel corpus construction from public sources}
\subsection{Data Crawling}
An essential component of any machine translation system is the parallel corpus. A good system requires a parallel corpus with a substantial  number  of  qualified  sentence  pairs. There are various projects building English-Vietnamese corpora for specific tasks such as word-sense disambiguation~\cite{Dinh:2002:BTC:1118794.1118801} \cite{Dien:2003:PEB:1118905.1118921}, VLSP project~\cite{22:VLSP}, web mining \cite{conf/rivf/DangH07}, etc. EVBCorpus \cite{Ngo:2013} is a multi-layered English-Vietnamese Bilingual Corpus containing over 10,000,000 words. 

However since corpora such as EVBCorpus or VLSP are not openly published, we first needed to build a high-quality large-scale English - Vietnamese  parallel corpus. We developed a web crawler  to collect English - Vietnamese sentences from 1,500 movie subtitles from the Internet. We also use the TED Talk subtitles collected in \cite{24:cettoloEtAl:EAMT2012}. 

\begin{figure}[h]
\begin{minipage}{0.45\textwidth}
\centering
\includegraphics[scale=0.4]{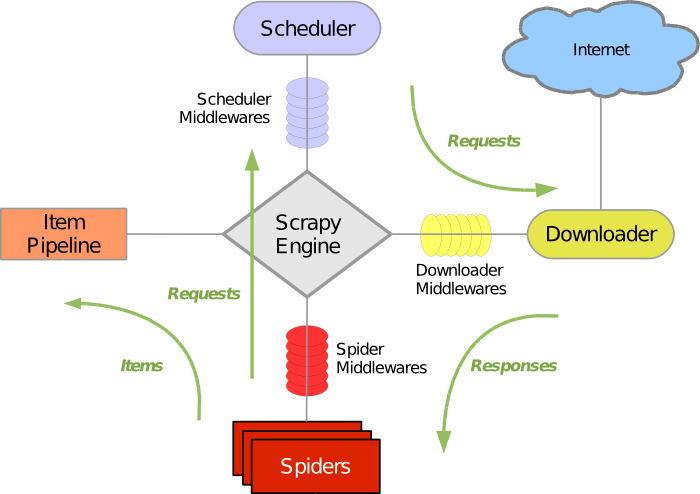}
\caption{Scrapy-based crawler engine architecture. Source: http://scrapy.org}
\label{tf_scrapy}
\end{minipage}~
\begin{minipage}{.45\textwidth}
\begin{tabular}{c|cc}
dataset & number of lines & number of tokens \\
\hline
train.en & 886224 & 10151378 \\
train.vi & 886224 & 11454886 \\
tst2012.en & 1553 & 28723\\
tst2012.vi & 1553 & 34345\\
tst2013.en & 1268 & 27317\\
tst2013.vi & 1268 & 33764\\
tst2015.en & 1080 & 21332\\
tst2015.vi & 1080 & 25341\\
\end{tabular}
\captionof{table}{Details of experiment dataset.}
\label{table:data}
\end{minipage}
\end{figure}

We use the Scrapy framework \cite{Scrapy:2016} to build our own crawler which consists of the following components (Figure~\ref{tf_scrapy})

\begin{description}[style=multiline,leftmargin=3cm,font=\normalfont]
\item[\rm Scrapy Engine] This component has the responsibility to control the data flow between components for coordination and triggering system events. 
\item[Scheduler] The Scheduler implements strategies to order URL crawling requests received from the Engine. 
\item[Downloader] The Downloader is responsible for fetching web pages and return crawled data to the Scrapy Engine.
\item[Spider] Spiders is responsible to parse responses and extract items from them or to perform additional requests to follow. We had to write our-own spiders classes to extract parallel English-Vietnamese sentences from HTML contents, based on CSS selectors and XPath expressions. 
\item[Item pipelines] The Item Pipeline processes the items after being extracted by the spiders. We defined our pipeline module to store scrapped items into a Mongo database instance. 
\item[Downloader middleware] Downloader middlewares hook between the Engine and the Downloader to intercept requests and responses. We had to write several downloader middlewares to rotate proxies, user-agents in order to improve our crawlers stability.  

\item[Spider middleware] Spider middlewares process spider input and output.
\end{description}
\vspace{0.5cm}
Built on Scrapy, we do not have to implement all the above components. We instead implemented only Spider, Item pipelines, and Downloader middlewares. Our web crawler collected around 1.2 millions parallel English - Vietnamse sentences.

\subsection{Data Cleaning and Preprocessing}
The following steps were conducted to clean the dataset:
\begin{itemize}
\item Detecting and removing incomplete translations: A big part of our dataset is movie subtitles, .where we found many partially translated examples. In order to detect and remove such subtitles, we use Princeton WordNet \cite{Miller:1995:WLD:219717.219748} to extract an English vocabulary. We then scan each subtitle for tokens found in the vocabulary. If a half of all tokens match this criteria, the subtitle is considered untranslated. We also use $langdetect$ package \footnote{https://pypi.python.org/pypi/langdetect} to filter out sentences which are not in Vietnamese. Manual observation on a random subset of removed subtitles shows that this heuristic filter works sufficiently for our purpose.
\item Removing low quality translations: There are many of low quality translations in our collected data, which we had to remove manually.
\end{itemize}
After filtering we obtained 886,224 sentences pairs for training. We use tst2012 for validation; tst2013, tst2015 for testing; all the thress are from IWSLT as provided in \cite{24:cettoloEtAl:EAMT2012}. The sizes of the datasets are shown in Table \ref{table:data}.

Following\cite{Phan:2017} we only use subword for our experiments. In particular we created a shared subword code file using Byte Pair Encoding (BPE) \cite{Sennrich:2016} using 32,000 merge operations. This shared subword code file was then used to transform the train, validation and test datasets to sub-words with a vocabulary size of approximately 20,000.
\section{Experiments and discussions}

\subsection{Overview of Training Configurations}
For authenticity the experiments with each model are performed on original software provided by the authors. Specifically Sequence to Sequence RNN experiments are performed using \cite{Luong:2017}, the Transformer experiments are performed using Tensor2Tensor (T2T) software \cite{Vaswani:2018} and the experiments on ConvS2S are performed using Facebook AI Research Sequence-to-Sequence Toolkit \cite{Gehring:2017}. 

Training is performed on a single Nvidia Geforce Titan X. We run each experiment 3 times with random initializations and save one model checkpoint every 1000 steps. The checkpoint for reporting results is selected based on BLEU score for the validation set. We train and report the model's performance at the maximum of 64th epoch due to our computing resource constraints. For the sake of brevity, we only report mean BLEU on our result tables.

In all our experiments, there are some common terms in all the models, which are specified as follow:
\begin{itemize}
\item \textbf{Maximum input length} (\texttt{max\_length}): specifies the maximum length of a sentence in tokens (sub-words in our case). Sentences longer than \texttt{max\_length} are either excluded from the training (T2T) or cut to match the \texttt{max\_length} (RNN). Lowering \texttt{max\_length} allows us to use a higher batch size and/or bigger model but biases the translation towards shorter sentences. Since $99\%$ of the training sentences are not longer than 70, we set \texttt{max\_length} to 70.

\item \textbf{Batch size} (\texttt{batch\_size}) For T2T  \texttt{batch\_size} is the approximate number of tokens (subwords) consumed in one training step, while for ConvS2S and RNN \texttt{batch\_size} is the number of sentence pairs consumed in one training step. Hence for consistency we define \texttt{batch\_size} as the approximate average number of tokens consumed in one training step. In fact the number of tokens in a sentence is the maximum of source and target subwords from the pair of training sentences. During training this allows us to put as many training tokens per batch as possible while ensuring that batches with long sentences still fit in GPU memory. In contrast, if we fixed the number of sentence pairs in a training batch, the model can run out of memory if a batch has many long sentences.
\item \textbf{Training epoch} is one complete pass through the whole training set. The number of training steps can be converted to epochs by multiplying by the batch size and dividing by the number of subwords in the training data.
\item \textbf{Model size} is number of trainable parameters of each model. Because of the difference in model structures, it is almost certain that two models with the same model size will not have the same training time.
\end{itemize}

Human judgment is always the best evaluation of machine translation systems; but in practice it is prohibitively expensive in time and resources. Therefore automatic scoring systems that evaluate machine translations against standard human translations are more commonly used. The most popular automatic metric in use is undoubtedly the BLEU score \cite{Papineni:2002}. BLEU has a high correlation with human judgments of quality and is easy to compute. Even though there are some acknowledged problems with BLEU and others better-performing metrics \cite{Bojar:2017}, we still stick to BLEU for its simplicity. In this work, we use the case-insensitive
sacréBLEU \footnote{https://github.com/awslabs/sockeye/tree/master/contrib/sacrebleu} version which uses a fixed tokenization.

\subsection{Sequence-to-Sequence RNN}
Based on previous literature in \cite{Phan:2017}, \cite{Britz:2017}, we build a baseline which is set reasonably large for our dataset.
We use LSTM \cite{Hochreiter:1996} for the two models as suggested in \cite{Britz:2017}. The embedding dimension in the baseline model is set to be equal to the number of cells in each layer. We use two layers of 1024-unit LSTMs for both the encoder and the decoder, whereas the encoder's first layer is a bi-directional LSTM network and each layer is equipped with a residual connection and a dropout of 0.15 is applied to the input of each cell.
We use Stochastic Gradient Descent (SGD) as the optimization algorithm with the batch size set to approximately 1280 tokens per step. The learning rate is set to 1.0; after 10 epochs we begin to halve the learning rate every single epoch. To prevent gradient explosion we enforce a hard constraint on the norm of the gradient by scaling it when its norm exceeds a threshold. In our two models the threshold is set to 5.0. For each training batch we compute $s = ||g||_2$ where $g$ is the gradient divided by the batch size. If $s > 5.0$, we set $g = \frac{5g}{s}$.

\begin{table}[h!]
\begin{center}
\caption{\small The \textit{baseline} system's performance with approximate 98 millions parameters}
\begin{tabular}{c|>{\centering\arraybackslash}m{1.5cm}>{\centering\arraybackslash}m{1.5cm}>{\centering\arraybackslash}m{2.5cm}|ccc}
task & batch size & training epochs & training time (days) & tst2012 & tst2013 & tst2015 \\
\hline
\textit{En-Vi} & 1300 & 32 & 3.5 & 34.77 & 37.00 & 31.03 \\
\textit{Vi-En} & 1300 & 20 & 2.5 & 35.21 & 38.83 & 30.29 \\
\hline
\end{tabular}
\end{center}
\end{table}

\subsection{Convolution Sequence to Sequence}
\begin{table}

\caption{\small The hyper-parameters set of ConvS2S model}
\label{convs2s:models_config}
\begin{center}
\begin{adjustbox}{max width=\textwidth}
\begin{small}
\begin{tabular}{c>{\centering\arraybackslash}m{2.8cm}>{\centering\arraybackslash}m{2.8cm}>{\centering\arraybackslash}m{1cm}cc|>{\centering\arraybackslash}m{1.5cm}>{\centering\arraybackslash}m{1cm}}
& encoder & decoder & emb size & $lr$ & $p_{drop}$ & training times (hours) & params $\times 10^6$ \\
\hline
$B_{base}$ & $4 \times [256 \times (3\times 3)]$ & $3 \times [256 \times (3\times 3)]$ & 256 & 0.5 & 0.1 & 4 & 10 \\
\hline
$B_1$ & $4 \times [512 \times (3\times 3)] \newline 2 \times [1024 \times (3\times 3)]  \newline 1 \times [2048 \times (1\times 1)]$ & $4 \times [512 \times (3\times 3)] \newline 2 \times [1024 \times (3\times 3)]  \newline 1 \times [2048 \times (1\times 1)]$ & 384 & 0.5 & 0.15 & 20 & 62 \\
\hline
$B_2$ &  $9 \times [512 \times (3 \times 3)] \newline 4 \times [1024 \times (3 \times 3)] \newline 2 \times [2048 \times (1 \times 1)]$ & $9 \times [512 \times (3 \times 3)] \newline 4 \times [1024 \times (3 \times 3)] \newline 2 \times [2048 \times (1 \times 1)]$ & 768 (512) & 0.5 & 0.15 & 48 & 144 \\ 
\hline
$B_3$ & $8 \times [512 \times (3 \times 3)] \newline 4 \times [1024 \times (3 \times 3)] \newline 2 \times [2048 \times (1 \times 1)] \newline 1 \times [4096 \times (1 \times 1)]$ & $8 \times [512 \times (3 \times 3)] \newline 4 \times [1024 \times (3 \times 3)] \newline 2 \times [2048 \times (1 \times 1)] \newline 1 \times [4096 \times (1 \times 1)]$ & 768 & 0.5 & 0.15 & 78 & 199 \\
\hline
\end{tabular}
\end{small}
\end{adjustbox}
\end{center}
\end{table}

We introduce four different models for each direction of translation. The hyper-parameters for each experiment are shown in Table \ref{convs2s:models_config}:
\begin{itemize}
\item We used Nesterov's accelerate gradient (NAG) \cite{Sutskever:2013} with a fixed learning rate: 0.25 for the $B_{base}$ model and 0.5 for the rests. After a certain number of epochs, we force-anneal the learning rate (\texttt{lr}) by a \texttt{lr\_shrink} factor: \texttt{lr\_new} = \texttt{lr} * \texttt{lr\_shrink}. We start annealing the learning rate at the 24th epoch with a width \texttt{lr\_shrink} of 0.1 for $B_{base}$ model and at the 50th epoch with the width \texttt{lr\_shrink} set to 0.2 for the rest. Once the learning rate falls below $10^{-5}$ we stop the training process.

\item In the $B_{2}$ model, we use embedding size of 768 for all internal embeddings except the decoder output embedding (pre-softmax layer) which is set to 512. 
\item The effective context size of $B_{base}$, $B_1$, $B_2$ and $B_3$ are 9, 13, 27 and 25, respectively.

\item We apply label smoothing of $\epsilon_{ls} = 0.1$ for all 4 models. This makes training perplexity fluctuate in a small interval but improves accuracy and BLEU score \cite{Vaswani:2017}.

\end{itemize}

We use cross-validation's BLEU score to decide which checkpoint to select for evaluation: the $B_{base}$ model is evaluated after 32  training epochs, the rest are evaluated after 64 training epochs. We found that the best perplexity of the validation dataset does not correspond to the best BLEU score on the test set but the BLEU score on the validation dataset does.

\begin{table}[h!]
\caption{ BLEU score of English-Vietnamese and Vietnamese-English translation task on tst2012, tst2013, tst2015 of $B_{base}$, $B_1$, $B_2$ and $B_3$ model using beam-search with length penalty set to 1, $b$ is the beam size}
\label{convs2s-envi}
\begin{center}
\begin{adjustbox}{max width=\textwidth}
\begin{tabular}{|l|ccc|ccc|}
\hline
model & \multicolumn{3}{c|}{English-Vietnamese} & \multicolumn{3}{c|}{Vietnamese-English} \\
& tst2012 & tst2013 & tst2015 & tst2012 & tst2013 & tst2015 \\
\hline
$B_{base}$, $b = 1$ & 24.31 & 25.34 & 23.98 & 25.18 & 27.76 & 23.89 \\
$B_{base}$, $b = 2$ & 26.40 & 28.02 & 26.56 & 26.09 & 27.42 & 24.89\\
$B_{base}$, $b = 5$ & 26.92 & \textbf{28.75} & 27.86 & 26.74 & 28.60 & 25.36  \\ 
$B_{base}$, $b = 10$ & 27.09 & 28.64 & 27.87 & \textbf{26.97} & \textbf{29.21} & 25.59\\ 
$B_{base}$, $b = 20$ & 27.08 & 28.66 & 28.09 & 26.86 & 29.46 & \textbf{25.61}  \\
$B_{base}$, $b = 100$ & \textbf{27.22} & 28.55 & \textbf{28.18} & 26.83 & 29.31 & 25.60\\
\hline
$B_1$ , $b = 1$ & 25.29 & 27.01 & 24.99 & 26.08 & 28.91 & 24.57\\
$B_1$ , $b = 2$ & 27.97 & 29.01 & 27.15 & 27.67 & 30.07 & 26.38\\ 
$B_1$ , $b = 5$ & 29.39 & 31.77 & 28.88 & 28.35 & 31.24 & 27.16\\ 
$B_1$ , $b = 10$ & 29.86 & \textbf{32.26} & 29.31 & 28.40 & 31.63 & 27.32\\ 
$B_1$ , $b = 20$ & \textbf{29.94} & 32.25 & 29.41 & 28.44 & 31.86 & 27.24\\ 
$B_1$ , $b = 100$ & 29.84 & 32.15 & \textbf{29.75} & \textbf{28.58} & \textbf{31.87} & \textbf{27.39}\\
\hline
$B_2$ , $b = 1$ & 34.87 & 36.57 & 29.13 & 37.90 & 39.11 & 28.33\\
$B_2$ , $b = 2$ & 36.36 & 37.61 & 30.42 & 39.85 & 41.78 & 29.99\\
$B_2$ , $b = 5$ & 37.20 & \textbf{38.53} & 31.10 & 41.07 & 42.85 & 30.52\\
$B_2$ , $b = 10$ & 37.19 & 38.48 & 31.23 & 41.19 & 43.03 & 30.60\\
$B_2$ , $b = 20$ & 37.36 & 38.36 & 31.25 & 41.44 & 43.32 & \textbf{30.74}\\
$B_2$ , $b = 100$ & \textbf{37.49} & 38.42 & \textbf{31.42} & \textbf{41.49} & \textbf{43.36} & 30.71\\
\hline
$B_3$ , $b = 1$ & 40.38 & 40.81 & 31.40 & 42.63 & 44.17 & 32.68\\
$B_3$ , $b = 2$ & 41.87 & 42.62 & 32.04 & 43.09 & 45.61 & 33.41\\
$B_3$ , $b = 5$ & 42.32 & 42.49 & 33.56 & 44.48 & 46.13 & 34.01\\
$B_3$ , $b = 10$ & 42.40 & 43.51 & \textbf{33.50} & 44.32 & 46.31 & \textbf{34.11}\\
$B_3$ , $b = 20$ & \textbf{42.51 }& 43.56 & 33.39 & 44.31 & 46.42 & 34.10\\
$B_3$ , $b = 100$ & 42.26 & \textbf{43.60} & 33.48 & \textbf{44.52}& \textbf{46.45} & 33.99\\
\hline
\end{tabular}
\label{convs2s-results}
\end{adjustbox}
\end{center}
\end{table}

We did not observe over-fitting with the large number of parameters from the results in Table \ref{convs2s-results}, that suggests the training data is fairly good and the model's dropout probability is suitable. With the hypothesize that the models' beam size and length penalty parameters are independence, we found that all the model's BLEU scores are improved a lots when beam size is increased from $b$ = 1 to $b$ = 10 and is only improved by a small margin (or even worst) when we keep increasing the beam size further. Because the decoding speed would slow down when we increase the beam size, we can conclude that the beam size of the ConvS2S models should set to 10.
\begin{figure}[h!]
	\begin{subfigure}{.5\textwidth}
	\pgfplotsset{width=0.94\textwidth,compat=1.14}
    \centering
	\begin{tikzpicture}
	\begin{axis}[
	xmin=1,
	xmax=101,
	xlabel= beam size ($b$),
	ylabel= decoding speed]
	\addplot[color=black, mark=*] coordinates {
	(1 , 2660)
	(2 , 3066)
	(5 , 2981)
	(10 , 2420)
	(20 , 1890)
	(100 , 554)
	};
    \addlegendentry{$B_{base}$}
	
    \addplot[color=green, mark=square*] coordinates {
	(1 , 1468)
	(2 , 1400)
	(5 , 1320)
	(10 , 1020)
	(20 ,  640)
	(100 , 190)
	};
	\addlegendentry{$B_1$}
    
    \addplot[color=blue, mark=triangle*] coordinates {
	(1 , 1260)
	(2 , 1180)
	(5 ,  960)
	(10 ,  690)
	(20 ,  430)
	(100 , 109)	
	};
	\addlegendentry{$B_2$}
    
    \addplot[color=red, mark=diamond*] coordinates {
	(1 , 1070)
	(2 , 890)
	(5 , 680)
	(10 , 490)
	(20 , 300)
	(100 , 60)	
	};
	\addlegendentry{$B_3$}
	\end{axis}
	\end{tikzpicture}
    \caption{}
	\end{subfigure}~
	\begin{subfigure}{.5\textwidth}
	\pgfplotsset{width=0.94\textwidth,compat=1.14}
    \centering
	\begin{tikzpicture}
	\begin{axis}[
	xmin=1,
	xmax=101,
	xlabel= beam size ($b$),
	ylabel= decoding speed]
	\addplot[color=black, mark=*] coordinates {
	(1 , 3500)
	(2 , 1170)
	(5 , 840)
	(10 , 410)
	(20 , 124)
	(100 , 35)
	};
    \addlegendentry{$C_{base}$}
	
    \addplot[color=green, mark=square*] coordinates {
	(1 , 890)
	(2 , 513)
	(5 , 224)
	(10 , 116)
	(20 ,  56)
	(100 , 12)
	};
	\addlegendentry{$C_1$}
    
    \addplot[color=blue, mark=triangle*] coordinates {
	(1 , 260)
	(2 , 137)
	(5 ,  85)
	(10 ,  29)
	(20 ,  13)
	(100 , 5)	
	};
	\addlegendentry{$C_2$}

	\end{axis}
	\end{tikzpicture}
    \caption{}
	\end{subfigure}
	\caption{\small The impact of beam size to the decoding speed of ConvS2S models (a) and Transformer models (b). The decoding speed of ConvS2S models is often higher and decrease slower when increase the beamsize than the decoding speed of Transformer models with same model size.}
\end{figure}
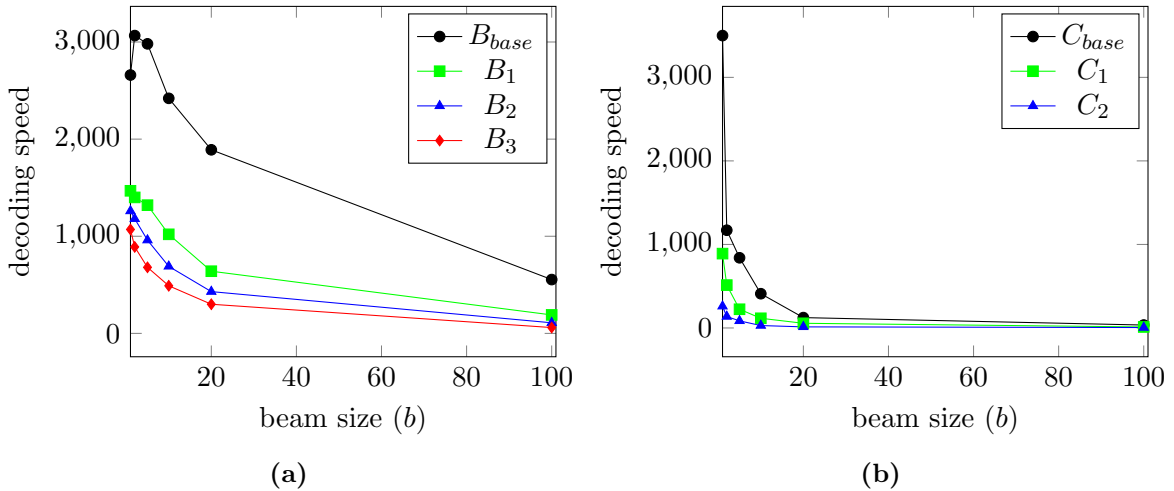

\subsection{Transformer}
In the Transformer architecture there are many hyper-parameters to be configured such as the number of layers in the encoder and the decoder, the number of attention heads or the size of the FFN weight matrix etc. In this work, we introduce three models based on their number of parameters. Each model's hyper-parameters are shown Table \ref{tf-hp}:
\begin{itemize}
\item We used the Adam optimizer with $\beta_1 = 0.9$, $\beta_2 = 0.98$ and $\epsilon = 10^{-9}$. The learning rate is varied over the training processes according to the following formula: $lr = d_{model}^{-0.5} \cdot min(step_{num} ^ {-0.5}, step_{num} ^ {-0.5} \cdot warmup_{steps}^ {-1.5})$, where \texttt{warmup\_steps} is set to 4000 from $C_{base}$ and \texttt{warmup\_steps} is set to 16000 for the rests.

\item In the course of training we found that if \texttt{batch\_size} is too big, the model can sometimes run out of GPU memory after a long training time. Therefore while $C_2$ can be trained with a \texttt{batch\_size}  of 3000 and the rest can be trained with a \texttt{batch\_size} larger than 6500, we recommend a batch size of 2048 for the $C_2$ and 4096 for the rests in order to keep the training stable. Another reason is our observation that the time to convergence does not change significantly once the batch size gets sufficiently large.
\item The learning rate are chosen based on \texttt{batch\_size}. Specifically, we set the learning rate to 0.0001 for the largest model with beam size of 2048 and scale it by $\sqrt{k}$ when multiplying the \texttt{batch\_size} by $k$.

\item We also observed that when the \texttt{batch\_size} is too small (i.e. $< 512$ for the biggest model), the model can only converge when the learning rate is smaller than 0.00005. Even then the model's BLEU is much lower than with a large \texttt{batch\_size}. This is due to the fact that the \textit{gradient noise scale} is proportional to the learning rate divided by the batch size. Thus, lowering the batch size increases the noise scale \cite{Smith:2017}. Therefore, we would rather reduce the model's complexity than reduce the batch size.

\item The performance of the $C_2$ model kept improving epoch by epoch and could potentially be better than reported. However due to resource constraints we report the model's performance at the $64^{th}$ checkpoint.
\end{itemize}

\begin{table}[h!]
\caption{Transformer hyper-parameters}
\label{tf-hp}
\begin{center}
\begin{tabular}{l|ccccp{0.8cm}|>{\centering\arraybackslash}m{1.5cm}>{\centering\arraybackslash}m{1.5cm}}

& $N$ & $d_{model}$ & $h$ & $d_{ff}$ & $p_{drop}$ & training time (hours) & params $\times 10^6$\\ 
\hline
$C_{base}$ & 2 & 256 & 4 & 1024 & 0.1 & 5 & 9M \\
\hline

$C_1$ & 6 & 512 & 16 & 2048 & 0.15 & 36 & 54M \\
\hline

$C_2$ & 8 & 1024 & 16 & 4096 & 0.15 & 72 & 197M\\

\hline
\end{tabular}
\end{center}
\end{table}

\begin{table}[h!]
\caption{BLEU score of English-Vietnamese translation task on tst2012, tst2013, tst2015 of $C_{base}$, $C_1$ and $C_2$ model, $b$ is the beam size with a default length penalty function}
\label{tf-results}
\begin{center}
\begin{tabular}{|l|ccc|ccc|}
\hline
model & \multicolumn{3}{c|}{English-Vietnamese} & \multicolumn{3}{c|}{Vietnamese-English}\\
\hline
$C_{base}$, $b =   1$ & 31.88 & 33.70 & 31.16 & 27.71 & 29.32 & 25.35\\
$C_{base}$, $b =   2$ & 31.99 & 34.44 & 31.54 & 28.63 & 30.02 & 25.94\\
$C_{base}$, $b =   5$ & \textbf{31.19} & \textbf{34.61} & \textbf{32.06} & \textbf{28.86} & \textbf{30.48} & 26.17\\ 
$C_{base}$, $b =  10$ & 31.93 & 34.44 & 31.79 & 28.74 & 30.01 & \textbf{26.28}\\ 
$C_{base}$, $b =  20$ & 31.86 & 34.18 & 30.79 & 28.55 & 30.95 & 26.14\\
$C_{base}$, $b = 100$ & 30.96 & 33.85 & 30.21 & 25.11 & 27.42 & 26.09\\
\hline
$C_1$, $b =   1$ & 36.85 & 39.88 & 33.62 & 33.31 & 35.71 & 29.58\\
$C_1$, $b =   2$ & 37.46 & \textbf{40.99} & 30.88 & 33.27 & 35.77 & 30.28\\ 
$C_1$, $b =   5$ & 37.61 & 40.88 & 33.48 & \textbf{33.32} & \textbf{36.06} & 30.33\\ 
$C_1$, $b =  10$ & 37.51 & 40.81 & 33.59 & 33.28 & 35.88 & \textbf{30.36}\\ 
$C_1$, $b =  20$ & \textbf{37.65} & 40.66 & \textbf{33.73} & 33.18 & 35.83 & 30.30\\ 
$C_1$, $b = 100$ & 37.43 & 40.02 & 33.31 & 32.62 & 34.99 & 30.17\\
\hline
$C_2$, $b =   1$ & 52.37 & 54.70 & 38.01 & 41.61 & 44.31 & 33.19\\
$C_2$, $b =   2$ & 52.89 & 55.31 & 38.81 & 43.16 & 45.41 & 33.83\\
$C_2$, $b =   5$ & 53.32 & \textbf{55.89} & \textbf{39.14} & \textbf{43.41} & 46.26& 33.94\\
$C_2$, $b =  10$ & \textbf{53.64} & 55.85 & 39.01 & 43.32 & 46.27 & 34.05\\
$C_2$, $b =  20$ & 53.50 & 55.76 & 39.05 & 43.10 & \textbf{46.49} & 34.20\\
$C_2$, $b = 100$ & 53.36 & 55.31 & 38.92 & 42.86 & 45.71 & \textbf{34.24}\\
\hline
\end{tabular}
\end{center}
\end{table}

\subsection{Length normalization for beam search}

\begin{figure}[h!]
	\pgfplotsset{width=7.5cm,compat=1.14}
    \centering
	\begin{tikzpicture}
	\begin{axis}[
	xmin=0,
	xmax=3.1,
	xlabel=$\alpha$,
	ylabel=BLEU score]
	\addplot[color=green, mark=*] coordinates {
		(0.1 , 36.21)
		(0.2 , 36.32)
		(0.3 , 36.41)
		(0.4 , 36.49)
		(0.5 , 36.64)
		(0.6 , 36.69)
		(0.7 , 36.78)
		(0.8 , 36.82)
        (0.9 , 36.94)
        (1.0 , 37.07)
        (1.1 , 37.11)
        (1.2 , 37.26)
        (1.3 , 37.28)
        (1.4 , 37.36)
        (1.5 , 37.39)
        (1.6 , 37.36)
        (1.7 , 37.37)
        (1.8 , 37.45)
        (1.9 , 37.45)
        (2.0 , 37.46)
        (2.1 , 37.42)
        (2.2 , 37.45)
        (2.3 , 37.46)
        (2.4 , 37.47)
        (2.5 , 37.47)
        (2.6 , 37.51)
        (2.7 , 37.52)
        (2.8 , 37.51)
        (2.9 , 37.54)
        (3.0 , 37.55)
	};
    \addlegendentry{$f_1$}

    \addplot[color=red, mark=square*] coordinates {
		(0.1 , 36.21)
		(0.2 , 36.3)
		(0.3 , 36.38)
		(0.4 , 36.59)
		(0.5 , 36.51)
		(0.6 , 36.76)
		(0.7 , 36.84)
		(0.8 , 37.03)
        (0.9 , 37.09)
        (1.0 , 37.23)
        (1.1 , 37.30)
        (1.2 , 37.30)
        (1.3 , 37.31)
        (1.4 , 37.33)
        (1.5 , 37.30)
        (1.6 , 37.41)
        (1.7 , 37.50)
        (1.8 , 37.52)
        (1.9 , 37.52)
        (2.0 , 37.50)
        (2.1 , 37.49)
        (2.2 , 37.48)
        (2.3 , 37.53)
        (2.4 , 37.52)
        (2.5 , 37.51)
        (2.6 , 37.48)
        (2.7 , 37.42)
        (2.8 , 37.41)
        (2.9 , 37.38)
        (3.0 , 37.25)
	};
	\addlegendentry{$f_2$}
	\end{axis}
	\end{tikzpicture}
    \caption{The effect of length penalty factor $\alpha$ on BLEU of $B_2$ on tst2012 with beam size fixed to 10.}
    \label{lenpen}
\end{figure}
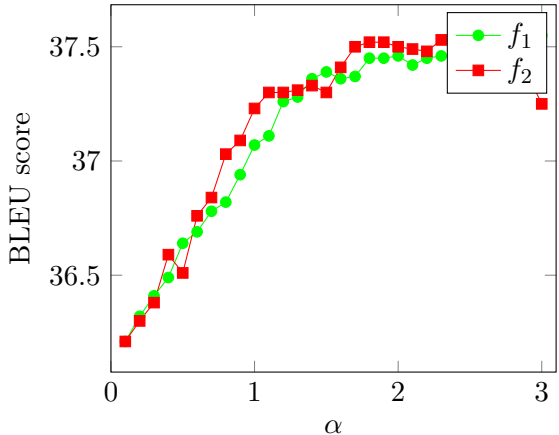

Beam search is a widespread technique in NMT, which finds the target sequence that maximizes some scoring function by a tree search. In the simplest case the score is the log probability of the target sequence. This simple scoring favors shorter sequences over longer ones on average since a negative log-probability is added at each decoding step.

Recently, length normalization \cite{Wu:2016} have been shown to improve decoding results for RNN based models. However, there is no guarantee that this strategy works well for other models. In this work, we experiment on two normalization functions described below:
\begin{align}
f_1 &= \frac{(5+|Y|)^\alpha}{6^\alpha} \\
f_2 &= (1+|Y|)^\alpha 
\end{align}
We found that the length penalty can help improve the model's performance up to 2 in BLEU scale. The length penalty should be chosen between 2.0 to 3.0.

\subsection{Ensembling}
Ensemble methods combine multiple individual methods to create a learning algorithm that is better than any of its individual parts \cite{Dietterich:2000}. They are widely used to boost machine learning models' performance \cite{Krogh:1994, Dietterich:2000}. In neural machine translation, the most popular ensemble method is checkpoint ensemble., in which the ensembled models are created by combining (averaging) multiple model checkpoints together \cite{Vaswani:2017,Wu:2016}. This method does not require training multiples model and the ensembled model has the size as same as the constituent models.

In many experiments the authors suggest to average checkpoints based on training time \cite{Vaswani:2017,1:Sutskever:2014:SSL:2969033.2969173}, which depends on hardware and hard to reproduce. In this work we experiment with checkpoint ensembling based on training epoch, which can be easily adapted to different platforms. 

\begin{table}[h!]
\caption{Effect of checkpoint ensembling ($n$ is number of checkpoint to be averaged) on the $B_2$ (a) and $C_1$ (b) model for English-Vietnamese translation task on tst2015}
\begin{center}
\begin{subfigure}{0.5\textwidth}
\caption{}
\begin{tabular}{c|cccc}
 $n$ & \multicolumn{4}{c}{interval (\% of a epoch)}\\
& 1.5 & 3 & 4.5 & 6 \\
\hline
8 & 32.26 & 32.74 & \textbf{32.82} & 32.68 \\
16 & 31.58 & 31.75 & 31.55 & 31.68\\
\end{tabular}
\end{subfigure}~
\begin{subfigure}{0.5\textwidth}
\caption{}
\begin{tabular}{c|cccc}
 $n$ & \multicolumn{4}{c}{interval (\% of a epoch)}\\
& 1.5 & 3 & 4.5 & 6 \\
\hline
8 & 30.63 & \textbf{30.90} & 30.77 & 30.81 \\
16 & 30.61 & 30.80 & 30.85 & 30.89 \\
\end{tabular}
\end{subfigure}
\end{center}
\end{table}

According to our experiments checkpoint ensembling always improves the model's performance. ConvS2S benefits the most (up to 7\% on the $B_2$ model) while ensembling has a smaller effect on the Transformer models (at most 5\% on the $C_3$ model) . We observed that taking 8 checkpoints for ensembling often yielded better results than 16 checkpoints. This also has the advantage that less time is spent on checkpoint saving.

Ensembling can also applied by training several new models starting form the same checkpoint . Each model is trained at a random position in the training data. In this setup, these models are semi-independent because they are rooted in the same source checkpoint. These semi-independent models can be averaged as described above, resulting a boost in the result, but in a smaller margin.

\section{Result and Empirical studies}
From the above experiments we observed that the training data is well correlated with the test data and training does not suffer from overfitting. However this makes it hard to tell if the model is general enough.

For new experiments we can always choose RNNs as a reliable base line model, that does not take much effort to achieve good results. The Transformer model has the highest converge speed while RNNs have the lowest converge speed. The Transformer model has showed its superiority in terms of achieving state-of-the-art results when given a suitable batch size and learning rate. Interestingly, even a very simple Transformer model with only 5 training hours can achieve a comparable score.

In our experiments we showed that a well-tuned beam search with length penalty is crucial, which can help the boost the model's score by 1.5 to 3 BLEU point (Table \ref{convs2s-results}, \ref{tf-results}, Figure \ref{lenpen} ). The most effective beam-size is 10 for ConvS2S and 5 for Transformer. The length penalty has a high impact on the final result, which should be set from 2 to 3.

Finally from our experiment results we compared our best performing hyper-parameter sets across all models and combined to a final model with the state-of-the-art results (Table \ref{combined-model} ). This shows that careful hyper-parameter tuning can greatly improve performance.

\begin{table}[h!]
\caption{\small Hyper-parameter settings for our final combined model for bidirectional English-Vietnamese translation}
\label{combined-model}
\begin{center}
\begin{tabular}{|r|l|}
\hline
Hyper-parameters & Value \\
\hline
$N$ & 8 \\
$d_{model}$ & 1024 \\
$h$ & 16 \\
$d_{ff}$ & 4096 \\ 
$p_{drop}$ & 0.15 \\
\texttt{batch\_size} & 2048 \\
\texttt{length\_penalty} & 1.5 \\
\texttt{beam\_size} & 5 \\
checkpoint to ensemble & 8 \\
ensemble interval & 3\% epoch \\
\hline
\end{tabular}
\end{center}
\end{table}

\begin{table}[h!]
\caption{\small BLEU results for our final combined model for bidirectional English-Vietnamese translation}
\label{combined-model}
\begin{center}
\begin{tabular}{c|ccc|ccc|}
& \multicolumn{3}{c|}{English-Vietnamese} & \multicolumn{3}{c|}{Vietnamese-English} \\ 
 & tst2012 & tst2013 & tst2015 & tst2012 & tst2013 & tst2015\\
\hline
Combined model & 55.04 & 56.88 & 40.01 & 46.36 & 49.23 & 35.81 \\
\hline
$C_2$ & 52.37 & 54.70 & 38.01 & 42.63 & 44.17 & 32.68\\
\hline
\end{tabular}
\end{center}
\end{table}

\section{Conclusion}
\label{sec:conclusions}
We conducted a broad range of experiments with the RNN sequence-to-sequence, ConvS2S and Transformer models for English-Vietnamese and Vietnamese-English translation, pointing out key factors to achieving state-of-the-art results. In particular we performed extensive exploration of hyper-parameters settings, which can be useful for other research works. In sum, our experiments took about 2,000 GPU hours.

We highlighted several important points: efficient use of batch size, the importance of beam search and length penalty, the importance of initial learning rate, the effectiveness of checkpoint ensembling, and the model's complexity. Along with these contribution we also make our dataset publicly available at (location withheld for review).

We hope our findings can help accelerate the pace of research on and application of English-Vietnamese and Vietnamese-English translation.

\section{Acknowledgement}
This research is funded by the Hanoi University of Science and Technology (HUST) under project number T2017-PC-078.

{\small
\bibliographystyle{IEEEtranS} \bibliography{main.bib}
 }

\end{document}